\documentclass[runningheads]{llncs}

\usepackage{eccv}

\usepackage{eccvabbrv}

\usepackage{graphicx}
\usepackage{booktabs}
\usepackage{multirow}
\usepackage{bm}
\usepackage{soul}
\usepackage[accsupp]{axessibility}  
\graphicspath{ {Figures/} }

\usepackage{hyperref}

\usepackage{orcidlink}

\begin{document}

\title{UTrack: Multi-Object Tracking with Uncertain Detections} 


\author{Edgardo Solano-Carrillo\orcidlink{0000-0002-9914-4666} \and
Felix Sattler \orcidlink{https://orcid.org/0000-0001-8869-282X}\and
Antje Alex \and
Alexander Klein \orcidlink{0009-0004-2403-0455}\and
Bruno Pereira Costa \and
Angel Bueno Rodriguez \orcidlink{https://orcid.org/0000-0002-7265-5642} \and
Jannis Stoppe \orcidlink{https://orcid.org/0000-0003-2952-3422}}

\authorrunning{E. Solano-Carrillo et al.}

\institute{German Aerospace Center (DLR)\\
Institute for the Protection of Maritime Infrastructures\\
\email{\{edgardo.solanocarrillo,felix.sattler,antje.alex,alexander.klein,\\bruno.costa,angel.bueno,jannis.stoppe\}@dlr.de}}

\maketitle



\begin{abstract}
  The tracking-by-detection paradigm is the mainstream in multi-object tracking, associating tracks to the predictions of an object detector.
  Although exhibiting uncertainty through a confidence score, these predictions do not capture the entire variability of the inference process.
  For safety and security critical applications like autonomous driving, surveillance, \etc, knowing this predictive uncertainty is essential though.
  Therefore, we introduce, for the first time, a fast way to obtain the empirical predictive distribution during object detection and incorporate that knowledge in multi-object tracking.
  Our mechanism can easily be integrated into state-of-the-art trackers, enabling them to fully exploit the uncertainty in the detections.
  Additionally, novel association methods are introduced that leverage the proposed mechanism.
  We demonstrate the effectiveness of our contribution on a variety of benchmarks, such as MOT17, MOT20, DanceTrack, and KITTI.
  \keywords{Multi-object tracking \and Uncertainty estimation}  
  
\end{abstract}

\newcommand{\angel}[1]{\textcolor{blue}{#1}}

\section{Introduction}
The safe application of artificial intelligence (AI) requires reliable uncertainty quantification \cite{varshney2017safety,safety_critical2018,begoli2019need,SEONI2023107441,ABDAR2021243, YU2022104814}, especially in visual perception systems. Much effort has focused on estimating uncertainty in object detectors \cite{rev_obj_det,10028728}. However, automated decision-making under visual uncertainty is dynamic, involving not only detecting fuzzy objects but also tracking their identities over time. These tasks are essential for an AI system to establish spatio-temporal semantic relationships in an uncertain environment.

Unifying these tasks using an end-to-end, real-time framework has seen fragmented progress. On one hand, uncertainty quantification in deep learning models has traditionally involved predicting targets and associated uncertainties as separate objectives within the same model \cite{pearce2020uncertainty,Ovadia,SQR,gawlikowski2023survey,pearce,Malinin,Maddox,Sensoy,hafner2019,9001195}. Applying this to visual perception using probabilistic detectors \cite{aut_driv,9525313} incurs computational costs that render online predictions infeasible. Although improvements in inference speed exist \cite{Choi_2019_ICCV, Choi_2021_ICCV,8917494}, the resulting models do not match the performance of established real-time detectors. On the other hand, the latter detectors require delicate speed-accuracy tradeoffs. Popular examples such as YOLOv8\cite{Jocher_YOLO_by_Ultralytics_2023} and DETR\cite{detr} are carefully tailored to achieve such balance. So there is hardly room to add uncertainty prediction heads without altering this balance.

Furthermore, beyond achieving reliable real-time detections with uncertainty estimates, the subsequent challenge lies in effectively utilizing this information for multi-object tracking. Currently, the predominant method involves utilizing only the detections alongside a motion model \cite{wang2020towards} (typically a Kalman filter) to associate current detections with predicted states. This association commonly relies on computing the intersection over union (IoU) \cite{rezatofighi2019generalized} between predicted bounding boxes and ground truth objects. To our knowledge, there have been no attempts to integrate object detection uncertainty into the association process of multi-object trackers.

Here, we propose methods to address these gaps. Specifically:
\begin{itemize}
    \item We develop a fast way to extract the predictive distribution of real-time object detectors such as YOLOv8. This gives cheap localization uncertainty.
    \item We formalize the propagation of this detection uncertainty through the tracking association mechanism. Particularly, we define disambiguation methods for uncertain IoUs and inject correct Kalman filter's observation noise.
\end{itemize}
We demonstrate the effectiveness of our approach on popular multi-object tracking benchmarks. Notably, adding the extracted uncertainty to a baseline model's Kalman filter outperforms existing methods on the most complex dataset, among the family of state-of-the-art trackers considered.

\section{Related Work}
\label{sec:related work}

\paragraph{\textbf{Uncertainty quantification}.} As mentioned, there have been efforts to make object detectors self-aware through reliable uncertainty quantification \cite{lee2022localization,vadera2022uncertainty,oksuz2023towards}. Besides the challenges encountered with online deployment, the paradigm of modeling uncertainty with deep networks faces explainability issues \cite{rev_interpret,surv_explain,Rudin2019,d2022underspecification}. This has led to interest in building uncertainty models on top of black-box predictors\cite{Romano_2019,brando2020,SolanoPIM2021}. Our approach aligns with this idea: instead of modifying successful real-time object detectors, we measure their aleatoric (localization) uncertainty \emph{after} inference. A fast method to do this is introduced using YOLOv8.

\paragraph{\textbf{Multi-object tracking}.} We focus on the family of \emph{simple online and realtime tracking} (SORT) methods, first introduced in \cite{Bewley2016SORT}. They only require observations from an object detector, alongside a Kalman filter\cite{khodarahmi2023review} as a stochastic predictor of these observations, and the Hungarian algorithm\cite{mills2007dynamic} to match new observations to predictions. This methodology has evolved during the last decade to make tracking more robust. DeepSORT \cite{wojke2017simple} introduced an appearance feature extractor to reduce identity switches in tracks. ByteTrack \cite{zhang2022bytetrack} became popular by introducing a method to associate every detection box. StrongSORT \cite{strongSORT} improves DeepSORT from the object detection, feature embedding, and trajectory association perspectives. BoT-SORT\cite{aharon2022bot} significantly improves ByteTrack, by introducing camera motion compensation (CMC) and a re-ID module. OC-SORT \cite{cao2023observation} revisits SORT and proposes a re-update procedure in the Kalman filter to fix the noise accumulated during object occlusions. This is improved by Deep OC-SORT\cite{maggiolino2023deep} after incorporating appearance cues beyond simple heuristic models. Recently, SparseTrack \cite{sparsetrack} improved BoT-SORT by considering the pseudo-depth of objects in the images during the association process. We focus on the SORT family due to its simplicity and for providing state-of-the-art performance. This makes our proof of concept more appealing.

\begin{figure}[!tbp]
     \centering
     \begin{subfigure}[b]{0.7\textwidth}
     \captionsetup{font=normal}
         \centering
         \includegraphics[scale=0.32]{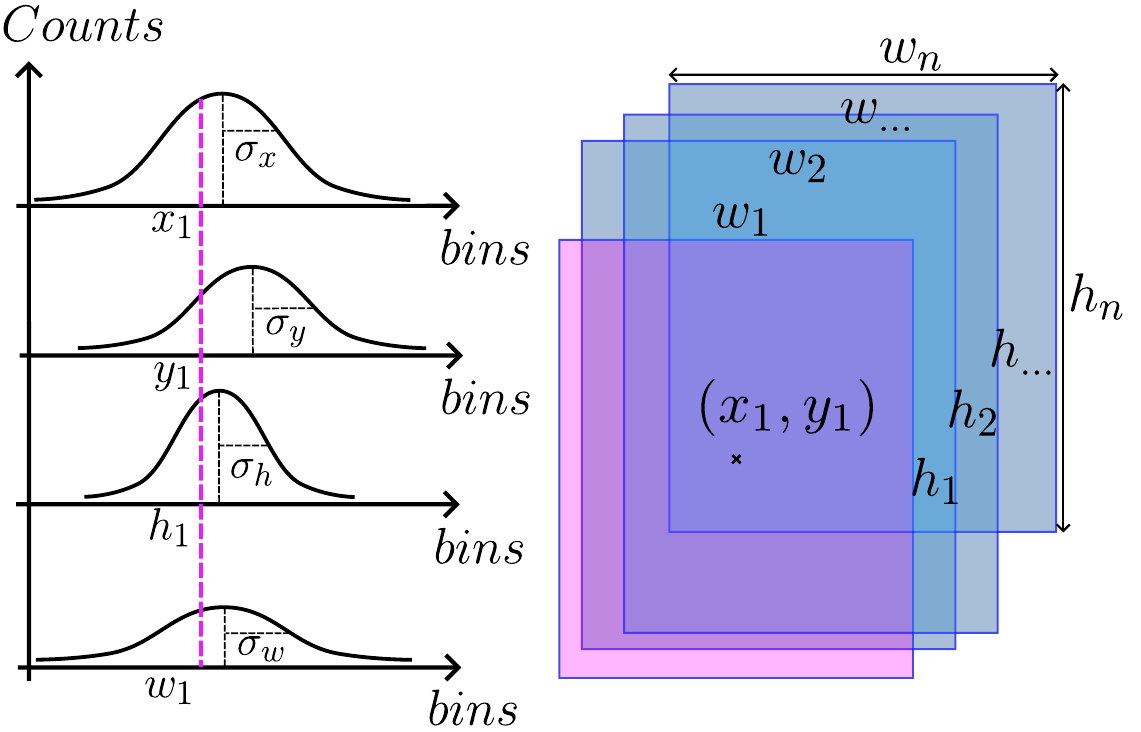}
         \vspace{4mm}
         \caption{}
         \label{fig:nms}
     \end{subfigure}
     \begin{subfigure}[b]{0.29\textwidth}
     \captionsetup{font=normal}
         \centering
         \includegraphics[scale=0.3]{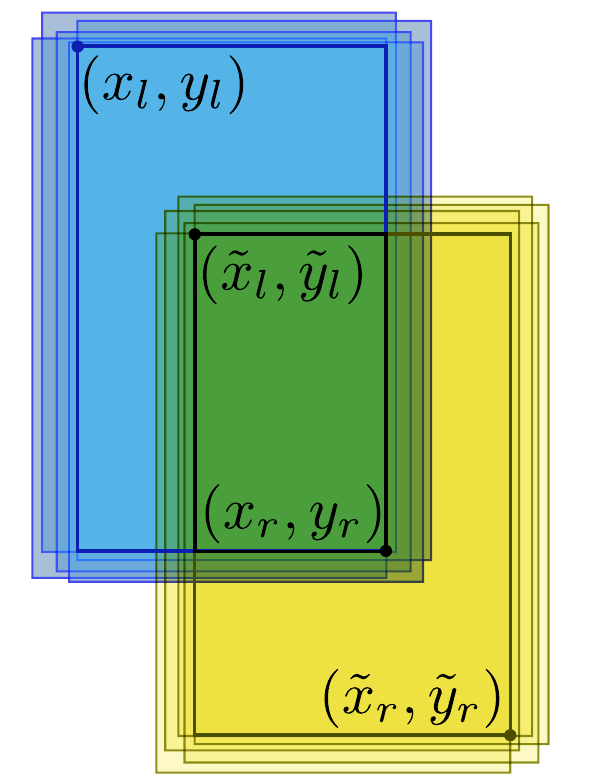}
         \vspace{4mm}
         \caption{}
         \label{fig:overlap}
     \end{subfigure}
     \caption{Aleatoric uncertainty in bounding-box regression and effect on IoU. (a) Object detection distribution before NMS. After NMS, only point estimates, \eg $(x_1, y_1, w_1, h_1)\,+$ confidence score, are provided. We measure $\sigma_a$ for $a\in\lbrace x,y,w,h\rbrace$. (b) Illustration of variance propagation to IoU. When the uncertainty is considered, the intersection area, $A_{\cap}$, of two bounding boxes becomes fuzzy (green region), as well as the union of areas.}
        \label{fig:NMS-IoU}
\end{figure}
\section{Tracking with uncertain detections}
Object detector models are trained to output $(x, y, w, h, s)$ per detected object. Here, $(x,y)$ are the pixel coordinates of the center of the bounding box enclosing the object. This box has width and height $(w, h)$, respectively. The model outputs a confidence score $s\in[0,1]$ correlated with the probability that the object belongs to an assigned class. For calibrated models \cite{guo2017calibration}, in which the observed accuracy of the predictor matches its mean confidence, the confidence score is then an estimate of such a probability.

Assuming that an object detector is calibrated, the confidence score only captures variability in the classification task, but not in the regression task. There are methods to calibrate confidence estimates of object detectors~\cite{kuppers2020multivariate}, which use information from the regression task. However, the resulting confidences are still point estimates. There are no variances that can be propagated to the quantities relevant for tracking, and that quantify the associated uncertainties. 

We empirically measure such variances, for the first time, and use them to quantify observation noise in the Kalman filter and to disambiguate bounding-box overlaps during multi-object tracking. The variance measurement is performed (\cref{fig:nms}) by obtaining the sample variance of all predicted bounding boxes (per object) during non-maximum-supression (NMS) \cite{gong2021review}.

\subsection{Uncertainty in IoU}
Since IoU is used in the tracker association process, we need an estimate of its relative error $\Delta := \delta \textrm{IoU}/\textrm{IoU}$. We use the relative error in the intersection area between two bounding boxes, $\delta A_{\cap}/A_{\cap}$. This area is $A_{\cap}=(x_r -\tilde{x}_l)(y_r -\tilde{y}_l)$, where $l$ and $r$ refer to the top left and bottom right coordinates of a bounding box (\cref{fig:overlap}). The addition rule for propagating relative errors then yields
\begin{equation}
 \dfrac{\delta A_{\cap}}{A_{\cap}}=\dfrac{\delta(x_r-\tilde{x}_l)}{x_r-\tilde{x}_l}+\dfrac{\delta(y_r-\tilde{y}_l)}{y_r-\tilde{y}_l}.
\end{equation}
This can be rewritten in terms of the center coordinates $(x,y)$ and box widths and heights $(w,h)$, by expressing $x_{l,r}=x\mp w/2$ and $y_{l,r}=y\mp h/2$, so
\begin{equation}
 \begin{split}
  \delta(x_r-\tilde{x}_l)&=\delta x-\delta\tilde{x}+\tfrac{1}{2}(\delta w+\delta\tilde{w}),\\
  \delta(y_r-\tilde{y}_l)&=\delta y-\delta\tilde{y}+\tfrac{1}{2}(\delta h+\delta\tilde{h}).
 \end{split}
\end{equation}
The uncertainty $\delta a$, with $a\in\lbrace x, y, w, h\rbrace$, is taken as the standard deviation $\sigma_{a}$ of the corresponding quantity, which we measure from the NMS procedure. This procedure allows us to attach an error (or uncertainty) $ \delta A_{\cap}/A_{\cap}$ to each IoU during tracking. We abuse notation and use $\delta A_{\cap}/A_{\cap}$ and $\Delta$ indistinguishably.

\subsection{IoU disambiguation}\label{subsec:disamb}
In each video frame, an object detection $d_j$ has to be associated with a track $T_i$, or form a new track. This is done by considering the IoU matrix $\textrm{IoU}_{ij}$. The rule followed by the Hungarian matching depends on which detection has the greatest overlap with a given track. Suppose that there are two detections, $d_j$ and $d_k$, which may have equal overlap with $T_i$, within error. This means that the intervals $\textrm{IoU}_{ij}\pm\Delta_{ij}$ and $\textrm{IoU}_{ik}\pm\Delta_{ik}$ overlap.  Which of the two detections is more likely to match that track? 

The above question cannot be answered without having more information. We therefore introduce the concept of a disambiguator:
\begin{definition}
 Let $\textrm{IoU}_{ij}$ be the IoU matrix referring to the box overlaps between track $T_i$ and detection $d_j$, with $\Delta_{ij}$ being the corresponding errors. Given a set $J=\lbrace j_1, j_2, \cdots,j_m\rbrace$ where the intervals $\textrm{IoU}_{ij_1}\pm \Delta_{ij_1}, \textrm{IoU}_{ij_2}\pm \Delta_{ij_2}, \cdots, \textrm{IoU}_{ij_m}\pm \Delta_{ij_m}$ overlap, a disambiguator $D_{ij}$ is a similarity matrix which defines a permutation $P(J)=\textrm{argsort}(D_{ij})_{j\in J}$ enforcing the correct ordering of IoUs in $J$.
\end{definition}
The intuition behind a disambiguator is simple. We cannot tell which is the greatest of all IoUs in a set of equally good ones (within error). Therefore, we need an external quantity that captures similarity between a detection and a track. This quantity breaks the ties, so the column order of IoU$_{i,j}$ in the set $J$ is obtained from the order of $D_{ij}$ in that set. We consider two specific examples next: disambiguating by phase and by bounding-box size.


\begin{figure}[tb]
  \centering
  \begin{subfigure}{0.35\linewidth}
  \centering
    \includegraphics[scale=0.32]{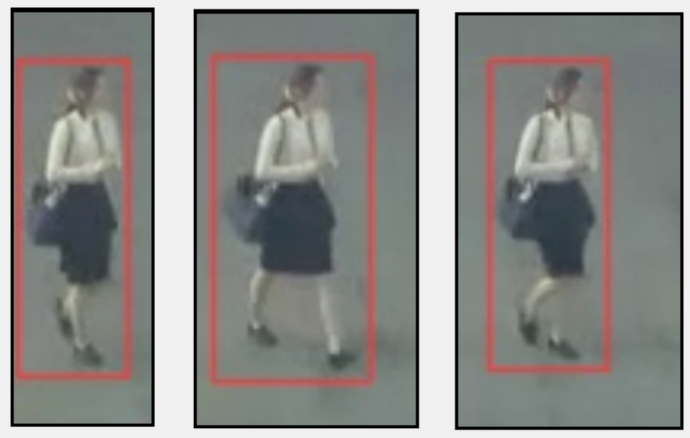} 
    \caption{Bounding-box oscillation.}
    \label{fig:phase-oscillate}
  \end{subfigure}
  \begin{subfigure}{0.6\linewidth}
  \centering
    \includegraphics[scale=0.22]{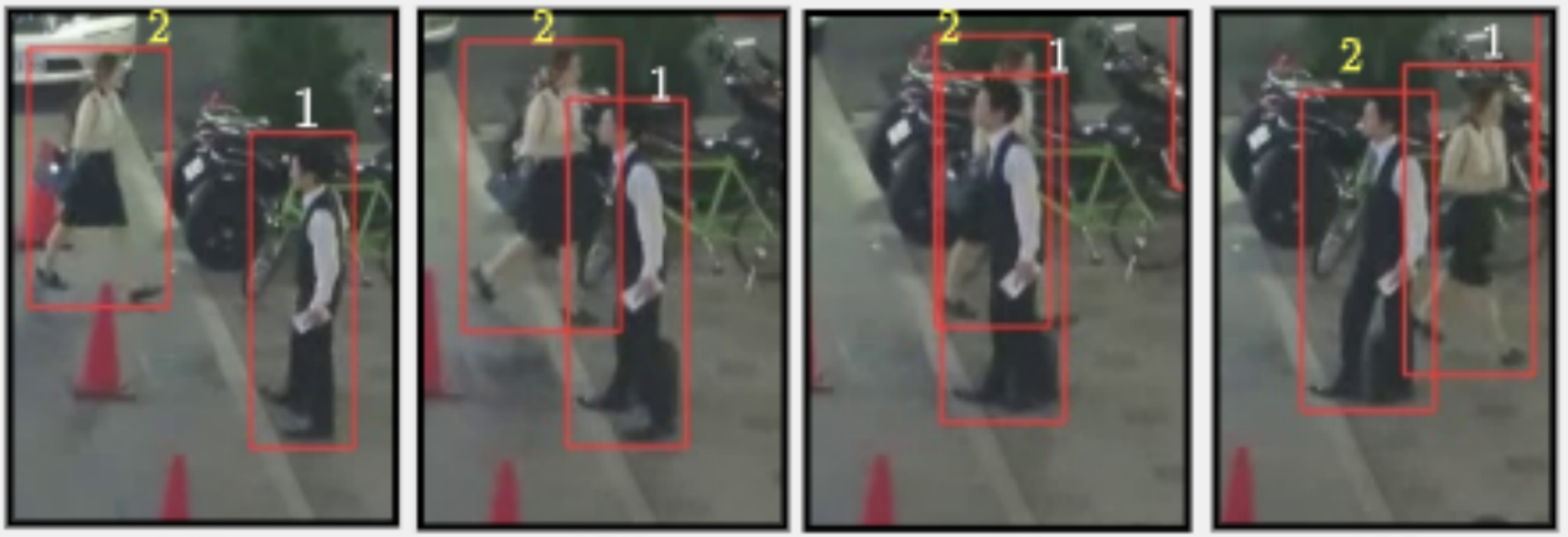}
    \caption{Awareness of oscillations  may prevent identity switches.
    }
    \label{fig:id-switch}
  \end{subfigure}
  \caption{Disambiguation of IoU using phase relationships (sequence from MOT17). By knowing that pedestrian 2 carries oscillations and pedestrian 1 does not, the object identities may become more robust to occlusion during tracking.} 
  \label{fig:short}
\end{figure}

\paragraph{\textbf{Disambiguating IoU by phase}.} We exploit, for the first time, the modeling of patterns arising from the internal motion of bounding boxes. These cannot be captured by the usual Kalman filter with constant velocity model. As shown in \cref{fig:phase-oscillate}, the width of the bounding boxes of pedestrians display a noisy quasi-periodic motion. We can leverage the associated patterns to aid the tracker association of pedestrians. The idea is illustrated in \cref{fig:id-switch}. The occlusion around the third snapshot easily leads a tracker to confuse the pedestrian identities. However, unlike pedestrian 1, pedestrian 2 is walking and therefore carries an oscillatory motion. A tracker disambiguating IoUs by phase will use this for matching, potentially reducing identity switches. 

In order to define the phase disambiguator, let us assume that we can predict the phase $\phi$ of the oscillatory motion of pedestrians, provided that there are indeed oscillations. We can quantify the presence of oscillatory behavior in pedestrians with the probability $p(X_{ij}^{\textrm{osc}}|X_{ij}^{\textrm{ped}})$ that the track $T_i$ and the detection $d_j$ are both oscillating bounding boxes, given that they are pedestrians. The closer $\phi_j$ is to $\phi_i$, the more likely should be the matching between $d_j$ and $T_i$. Therefore, the disambiguator is 
\begin{equation}\label{eq:pDij}
 D_{ij}^p = p(X_{ij}^{\textrm{osc}}|X_{ij}^{\textrm{ped}})\,C(\phi_i,\phi_j),
\end{equation}
where $C(\phi_i,\phi_j)=\tfrac{1}{2}[1+\cos(\phi_i-\phi_j)]$ is a phase similarity. $D_{ij}^p$ achieves its greatest value ($1$) for two bounding boxes oscillating in phase, and its lowest value ($0$) for two bounding boxes oscillating out-of-phase. 

Since $X_{ij}^{\textrm{osc}}=X_{i}^{\textrm{osc}}X_{j}^{\textrm{osc}}$ is a binary variable, we can easily estimate $p(X_{ij}^{\textrm{osc}})$ by counting the frequency of $1$'s in the binary matrix with elements $X_{ij}^{\textrm{osc}}$. Additionally, for oscillating boxes, we can write $p(X_{ij}^{\textrm{ped}}|X_{ij}^{\textrm{osc}})=p(X_{i}^{\textrm{ped}})p(X_{j}^{\textrm{ped}})X_{ij}^{\textrm{osc}}$. With this, we can estimate the $p$ in \cref{eq:pDij} by using Bayes theorem
\begin{equation}
 p(X_{ij}^{\textrm{osc}}|X_{ij}^{\textrm{ped}})=\dfrac{X_{ij}^{\textrm{osc}}\,p(X_{ij}^{\textrm{osc}})}{X_{ij}^{\textrm{osc}}p(X_{ij}^{\textrm{osc}})+(1-X_{ij}^{\textrm{osc}})(1-p(X_{ij}^{\textrm{osc}}))}.
\end{equation}
Note that this is independent of $p(X_{i}^{\textrm{ped}})$, so there is no need to calibrate the confidence score $s_i$ which would approximate $p(X_{i}^{\textrm{ped}})$ for the pedestrian class, if calibrated. This approach can be extended to other object classes if there is a good prior knowledge of the internal bounding-box motion.

\paragraph{Phase prediction.} How do we determine $\phi_i$, and hence $X_{i}^{\textrm{osc}}$? We achieve this by filling and rolling a buffer, for each track $T_i$, with the widths of the bounding boxes $w_i(t)$ --- and corresponding errors $\sigma_{w_i}(t)$. These are measured before the track is updated by the Kalman filter. We then define the binary variables $X_{i}^{\textrm{change}}$ and $X_{i}^{\textrm{back}}$. They indicate, respectively, whether there has been a significant change in widths since the last update, and whether the average $w_i$ (over the buffer) has gone back to the initial state. The criterion for change is that $w_{i}^{\textrm{max}}-w_{i}^{\textrm{min}} > 3 \bar{\sigma}_{w_i}$, with $\bar{\sigma}_{w_i}$ being the average of the width standard deviations, and $w_{i}^{\textrm{max/min}}$ being the width extremes, over the buffer. The criterion for returning to the initial state is that $|\bar{w}_i(t)-\bar{w}_i(t_0)|<\bar{\sigma}_{w_i}$. When both criteria are satisfied, they define an oscillation, so $X_{i}^{\textrm{osc}}=X_{i}^{\textrm{change}}\,X_{i}^{\textrm{back}}$. After each oscillation, the frequency is measured from the period $t-t_0$, and the initial state updated: $t_0\rightarrow t$. The frequency $\omega_i$ is obtained from this after smoothing, and the phase angle of track $T_i$ is
\begin{equation}
 \phi_i=(\omega_i\, t)\;\textrm{mod}\;\; 2\pi,
\end{equation}
where the time $t$ is measured from the frame number (frame) and the number of frames per second (FPS) as  $t=\textrm{frame}/\textrm{FPS}$.

\paragraph{\textbf{Disambiguating IoU by box sizes}.}
The phase is a dynamic feature which depends on the history of the tracks. For reference, we consider disambiguation by a similarity constructed from the area of the bounding boxes. This gives us statically-extracted features for IoU disambiguation --- \ie calculated without reference to frames different from the current one. The intuition is that, most of the time, the tracked objects have a consistent bounding box from one frame to the next one. This implies that it should be less likely to match track $T_i$ with detection $d_j$ if their bounding-box sizes differ ``too much''.
 
We can quantify this by standardizing the box areas, \ie $\tilde{a}_{T_i}=(a_{T_i} - \mu_T)/\sigma_T$ and $\tilde{a}_{d_j}=(a_{d_j} - \mu_d)/\sigma_d$, and defining the size disambiguator
\begin{equation}
 D_{ij}^{s} = \exp\bigl[-\tfrac{1}{2}(\tilde{a}_{T_i}-\tilde{a}_{d_j})^{2}\bigr],
\end{equation}
which behaves like the standard normal density, peaking around detected boxes with areas close to the tracked ones. Here, $(\mu_c,\sigma_c)$, with $c\in\lbrace T,d\rbrace$, are obtained from the bounding-box areas of all current tracks and detections, respectively.

\subsection{Kalman filtering with uncertain detections}\label{subsec:ukalman}
The state $\bm{s}_t:=(x_t, y_t, w_t, h_t)$ of track $T$ at time step $t$ is predicted by a Kalman filter. This state is projected from the variable $\bm{\bar{x}}_t:= (\bm{s}_t, \dot{\bm{s}}_t)$ evolving as
\begin{equation}
 \bm{\bar{x}}_t = \bm{F}_t\,\bm{\bar{x}}_{t-1} + \bm{w}_t,
\end{equation}
where $\bm{F}_t$ is the state transition model, usually implemented to give constant-velocity kinematics, and $\bm{w}_t\sim \mathcal{N}(0, \bm{Q}_t)$ is the process noise, with covariance $\bm{Q}_t$. The projection, $\bm{s}_t = \bm{H}_t\,\bm{\bar{x}}_t$,  is performed by a known (usually constant) matrix $\bm{H}_t$. For estimation, the Kalman filter is fed with observations $\bm{z}_t$ of $\bm{s}_t$, which are provided by the object detector. These are assumed to be related to the predicted variable as
\begin{equation}
 \bm{z}_t = \bm{H}_t\, \bm{\bar{x}}_t + \bm{o}_t,
\end{equation}
where $\bm{o}_t\sim \mathcal{N}(0, \bm{R}_t)$ is the observation noise, with covariance $\bm{R}_t$.

For multi-object tracking, it is customary to choose $\bm{Q}_t$ heuristically, by carefully fitting the relevant scales of the problem. As far as we know, either methods tracking in image space \cite{zhang2022bytetrack} or on the ground \cite{yi2024ucmctrack} also choose $\bm{R}_t$ parametrically; despite this quantity being related to \emph{measurements} of the state. We deviate from this practice and truly measure $\bm{R}_t:=\textrm{diag}(\sigma_{x}^{2}, \sigma_{y}^{2}, \sigma_{w}^{2}, \sigma_{h}^{2})_t$ during the NMS procedure, as indicated in \cref{fig:nms}. The resulting Kalman filter is called \texttt{U-Kalman}.

\subsection{Cascade associations}\label{subsec:cascade}
 The association of tracks $T_i$ to object detections $d_j$ is usually done in stages. First, the detections $d_j$ are ranked according to the values of a quantity $\mathcal{R}_j$ that is binned (\eg confidence score, pseudo-depth, \etc). Then, unmatched detections in the $r$-th bin of $\mathcal{R}$ are added to the detections in the $(r-1)$-th bin. The matching process then continues in a cascade fashion. ByteTrack \cite{zhang2022bytetrack} introduced this procedure, by taking $\mathcal{R}_j$ as the confidence scores, and binning the range in two halves (high and low scores). SparseTrack \cite{sparsetrack} uses the same procedure, but sub-divide each bin according to the pseudo-depth $\mathcal{R}_j$ of each detection.
 
 We implement a modular approach to cascade association, which allows us to compose the above and any other binning strategy. Unlike previous approaches, we add the possibility of using all unmatched detections from each bin (not just those from the first bin) in order to confirm new tracks. The effect of this is investigated in \cref{sec:results}. 
 
 \subsection{Camera motion compensation}\label{subsec:cmc}
 At every time step, an optimal Kalman filter predicts the tracked state $\bm{\bar{x}}_t$ and its covariance $\bm{P}_t=\textrm{cov}(\bm{x}_t-\bm{\bar{x}}_t)$. These have to be corrected if the camera is moving. The transformation law is dictated by the homography matrix
 \begin{equation*}
 \begin{pmatrix}
  \underset{\scriptscriptstyle 2\times 2}{\bm{H}_1} & \hspace{0.2cm}\underset{\scriptscriptstyle2\times 1}{\bm{h}_2}\\[0.4cm]
  \underset{\scriptscriptstyle1\times 2}{\bm{h}_3^T} & \hspace{0.2cm} \underset{\scriptscriptstyle1\times 1}{h_4}
 \end{pmatrix},
 \end{equation*}
 which relates how much the coordinate system $\Sigma_{t}$ of the camera at time $t$ has rotated and translated with respect to that $\Sigma_{t-1}$ at time $t-1$. Using this, the pixel coordinates of a point $p$ in the world coordinates transform as \cite{white2019homography}  
 
 \begin{equation}\label{eq:homography}
  \bm{p}_t = g(\bm{p}_{t-1}) = \dfrac{\bm{H}_1 \bm{p}_{t-1} + \bm{h}_2}{\bm{h}_3^T \bm{p}_{t-1} + h_4},
 \end{equation}
 during the perspective change. The velocity transforms as  $\bm{v}_t = \bm{G}(\bm{p}_{t-1})\bm{v}_{t-1}$, and the covariance as $\bm{K}_t=\bm{G}(\bm{\bar{p}}_{t-1})\bm{K}_{t-1} \bm{G}^T(\bm{\bar{p}}_{t-1})$, where $\bm{G}(\bm{p})=\partial g/\partial \bm{p}$. Note that the tracked variables, $\bm{\bar{x}}_t=(\bm{s}_t, \dot{\bm{s}}_t)$ and $\bm{P}_t$, are 8-dimensional. On the other hand, the transformation laws apply to 2-dimensional variables. So we need a method to apply the transformation laws to the relevant 2-dimensional sectors of the 8-dimensional space.
 
 BoT-SORT\cite{aharon2022bot} and SparseTrack\cite{sparsetrack},  simplify the problem by eliminating the non-linear nature of the transformation (brought about by the denominator in \cref{eq:homography}). This is done after extending the $\bm{H}_1$-component of the homography as $\bm{H}_{1;8\times 8} = \bm{1}_{4\times4}\otimes\bm{H}_1$. With this, the mean state of the Kalman filter is corrected as $\bm{\bar{x}}_t\rightarrow\bm{H}_{1;8\times 8}\;\bm{\bar{x}}_t + (\bm{h}_2,\bm{0},\bm{0},\bm{0})$ and the covariance as $\bm{P}_t\rightarrow\;\bm{H}_{1;8\times 8}\;\bm{P}_t\;\bm{H}_{1;8\times 8}^{T}$. Apart from the $\bm{h}_2$-shift in $\bm{\bar{x}}_t$, this affine approximation is nearly equivalent to transforming all 2-dimensional sectors by $\bm{H}_1$. As discussed in \cite{white2019homography}, this is only valid if the perspective change only involves rotations about the optical axis of the camera, and translations with this axis either pointing directly to the ground plane or perpendicular to it. In practice, this works well for distant scenes, small field of view of the camera, small camera rotations, among other cases.

 For high-speed camera motion, as possible in autonomous driving, the above approximation may fail. We improve on this by considering the exact transformation \cref{eq:homography} on $\bm{\bar{x}}_t=(\bm{s}_t, \dot{\bm{s}}_t)$, and an approximate transformation on $\bm{P}_t$. That is, we apply $g$ on the pixel representing the bounding-box center $(x,y)$, and on the top-left / bottom-right pixels of the bounding box. From the latter, the transformed $(w, h)$ can be obtained, and therefore $\bm{s}_t$. The computation of $\bm{G}$, for each case, allows us to transform $\dot{\bm{s}}_t$. Since $\lbrace w, h\rbrace$ relate to two pixels, not the single one involved in the definition of $\bm{G}$, the covariance cannot be easily transformed. For this reason, we extend to all other 2-dimensional sectors, the $\bm{G}_{xy}$ that correctly transforms the $x$-$y$ sector of the covariance. This is done by defining $\bm{G}_{xy;8\times 8} = \bm{1}_{4\times4}\otimes\bm{G}_{xy}$, so $\bm{P}_t\rightarrow\;\bm{G}_{xy;8\times 8}\;\bm{P}_t\;\bm{G}_{xy;8\times 8}^{T}$. We test this with the KITTI dataset for autonomous driving in \cref{sec:results}.


\section{Experiments}
In general, predictive uncertainty is classified as \emph{aleatoric} (related to observation noise) and \emph{epistemic} (related to model specification and  data distributional changes). Since we want to know the effect of aleatoric localization uncertaity of object detectors, the effect of the epistemic uncertainty has to be subtracted in the evaluation. A well-known method to probe the epistemic uncertainty is by model ensembling \cite{SolanoPIM2021}. This is however expensive, so we need a similar but cheaper method for computer vision.

Fortunately, the output of deep neural networks can be easily altered by adding relatively small perturbation to their inputs \cite{pixel-attack}. We therefore add stress to what the model specification expects and explore data distributional changes by running multiple versions of the same tracking configuration, each under random two-pixel attacks per video frame. The attack consists of randomly swapping two different pixels per image before it is given to the object detector.

We therefore deviate from the common practice in computer vision of comparing point estimates during ablation studies --- actually sometimes a lucky seed can lead to better reported performance \cite{Picard2021Torchmanual_seed3407IA}. In the following, we describe the experimental setup to evaluate this methodology. In \cref{subsec:datasets} we extract relevant features from the datasets used, in order to understand the results. In \cref{subsec:metrics} we describe the metrics relevant for multi-object tracking. Algorithmic implementation details\footnote{Code is publicly available at \url{https://github.com/DLR-MI/UTrack}.} are described in \cref{subsec:details}, and the results in \cref{sec:results}.

\subsection{Datasets}
\label{subsec:datasets}
We conduct experiments across four public benchmarks, each one with distinct characteristics regarding pedestrian densities, camera perspectives, and movements: MOT17 \cite{mot17}, MOT20 \cite{mot20}, DanceTrack \cite{dancetrack} and KITTI \cite{Geiger2013IJRR}. 

\paragraph{\textbf{MOT17}.} This dataset contains video sequences in city streets and inside shopping malls. These are acquired with moving or static cameras, overlooking the scene from high/medium/low positions, and distinct weather conditions. It poses various challenges such as occlusions, varying  densities $(\sim 8-70)$ of pedestrians per frame, and different camera angles. The varying complexity is then \emph{ideal to test our addition to the cascade association method} of \cref{subsec:cascade}.

\paragraph{\textbf{MOT20}.} This benchmark addresses the challenge of very crowded scenes. It contains sequences of a crowded square and people leaving the entrance of stadium, both at night time. Additionally, there are sequences of a pedestrian street scene in daylight and a crowded indoor train station. Compared with MOT17, the pedestrian densities are quite high $(\sim 62-246)$, with the scenes overlooked only from a high position. Nearly no (or sometimes very mild) camera motion is present.

\paragraph{\textbf{DanceTrack}.} The emphasized property of this dataset is that the multiple groups of dancers performing the choreographies have uniform appearance per group. Moreover, they execute diverse motions with complicated interaction patterns and frequent crossovers. This, combined with highly synchronized movements, makes it \emph{ideal to test our proposed phase disambiguation method} of \cref{subsec:disamb}. Moreover, DanceTrack has much larger volume compared with MOT datasets (with 10x more images and 10x more videos). So we use it to generalize the best of our observations from the validation across all datasets.

\paragraph{\textbf{KITTI}.} A standard station wagon, equipped with multiple sensors, collects data while driving on rural areas and highways. A dataset with annotated vehicles, cyclists, and pedestrians is then built. We only use the left color camera images for the visual pedestrian tracking task. KITTI has the lowest pedestrian density, with up to 30 pedestrians visible per frame. Unlike the other ones, there are usually large pixel displacements from frame to frame, brought about by fast camera motions. This makes it \emph{ideal to test our proposed camera motion compensation method}, as mentioned at the end of \cref{subsec:cmc}. 

\subsection{Metrics}\label{subsec:metrics}
Following recent practice, we use Higher Order Tracking Accuracy (HOTA)\cite{trackeval} as primary metric, combined with IDF1 \cite{Ristani2016PerformanceMA} and CLEAR \cite{bernardin2008evaluating} (MOTA). HOTA measures the multi-object tracking performance by balancing accurate detections (DetA) and accurate associations (AssA), taking into account object detector localization capabilities. IDF1 and AssA are used to evaluate how accurate the object identities are kept. DetA and MOTA evaluate detection performance. For all these metrics, a higher score represents better results.

\begin{figure}
\centering
	\begin{tabular}{ccc}
		& \hspace{7mm}  \textbf{ByteTrack} & \hspace{5mm} \textbf{BoT-SORT} \\
		\rotatebox{90}{
			\parbox{0.3\textwidth}{\centering{\textbf{MOT17}}}} &  \includegraphics[scale=0.3]{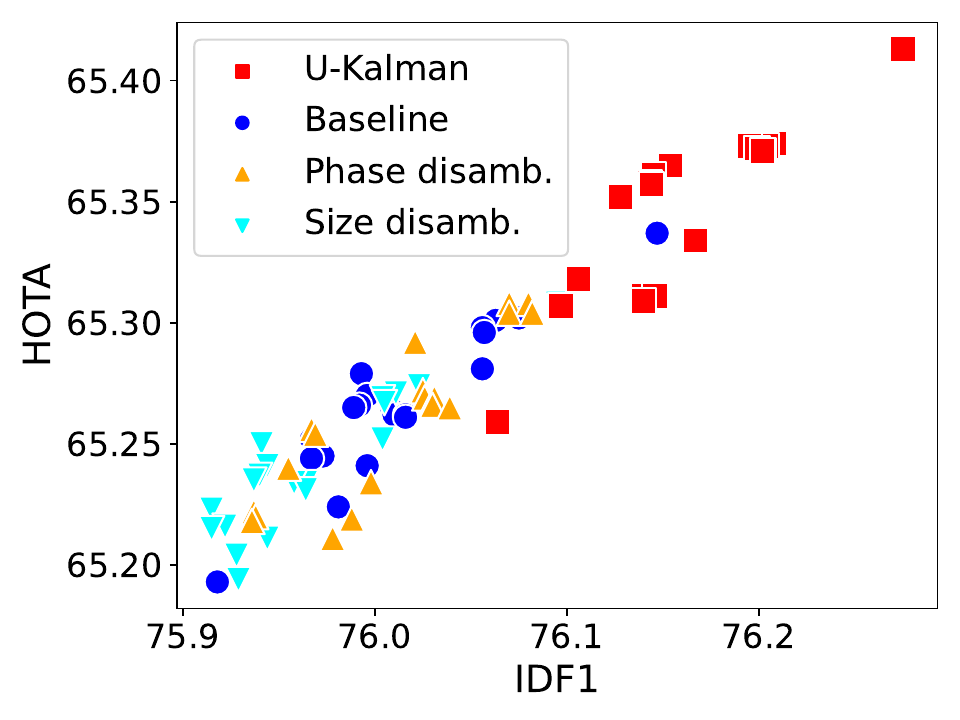}  & \includegraphics[scale=0.3]{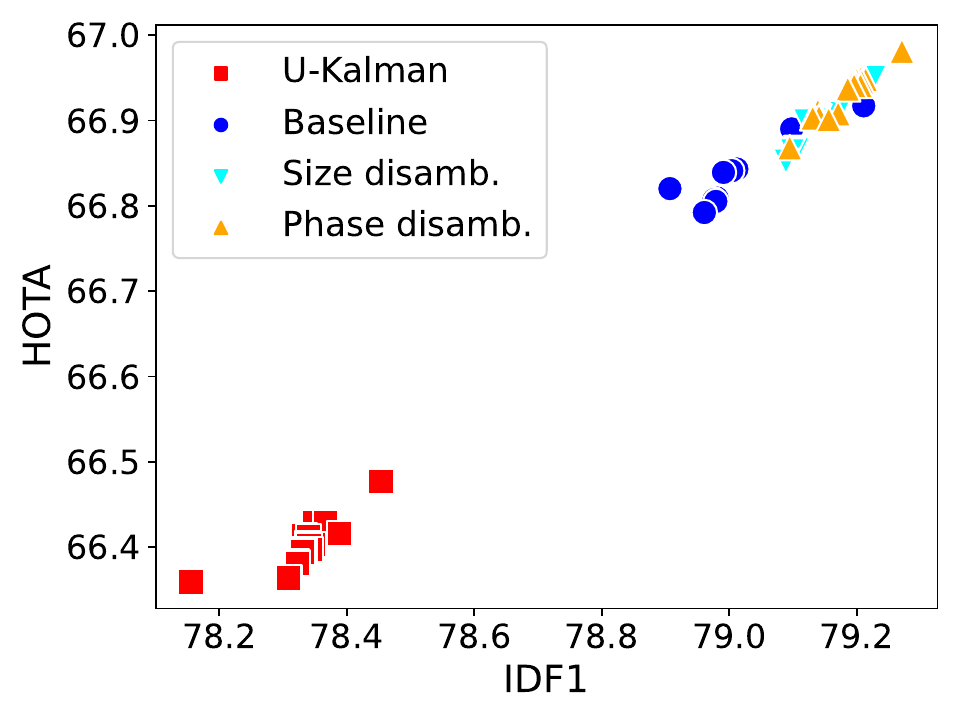} \\ 
		& \begin{subfigure}{0.2\linewidth}
		\caption{}
		\label{fig:byte_mot17} 
		\end{subfigure} & 
		\begin{subfigure}{0.05\linewidth}
		\caption{}
		\label{fig:botsort_mot17} 
		\end{subfigure}\\	
		\rotatebox{90}{
			\parbox{0.3\textwidth}{\centering{\textbf{MOT20}}}} &  \includegraphics[scale=0.3]{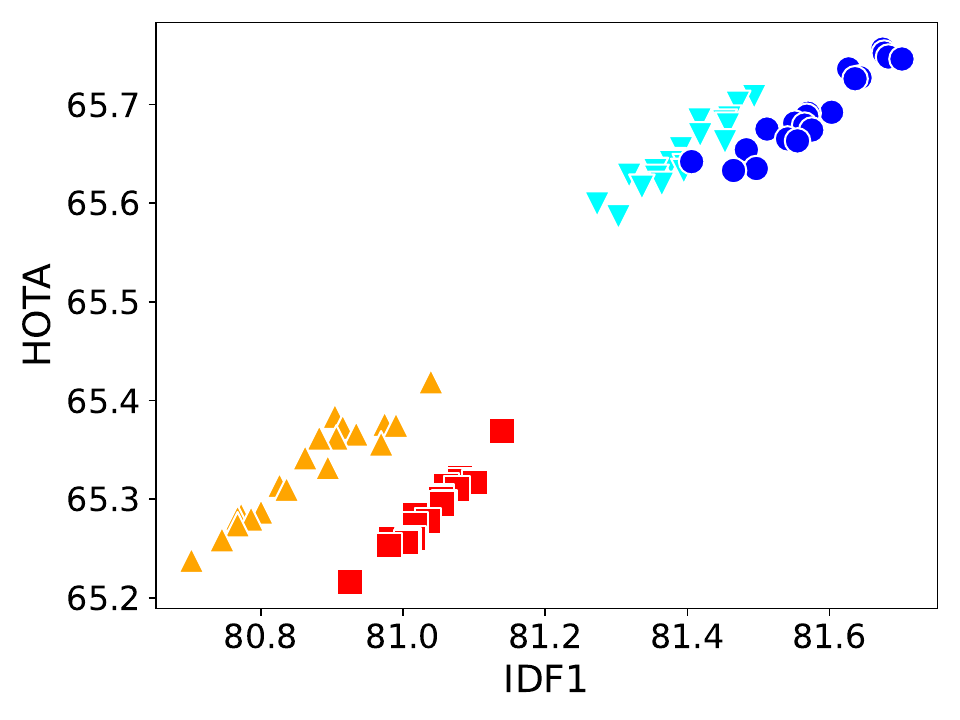}  &  \includegraphics[scale=0.3]{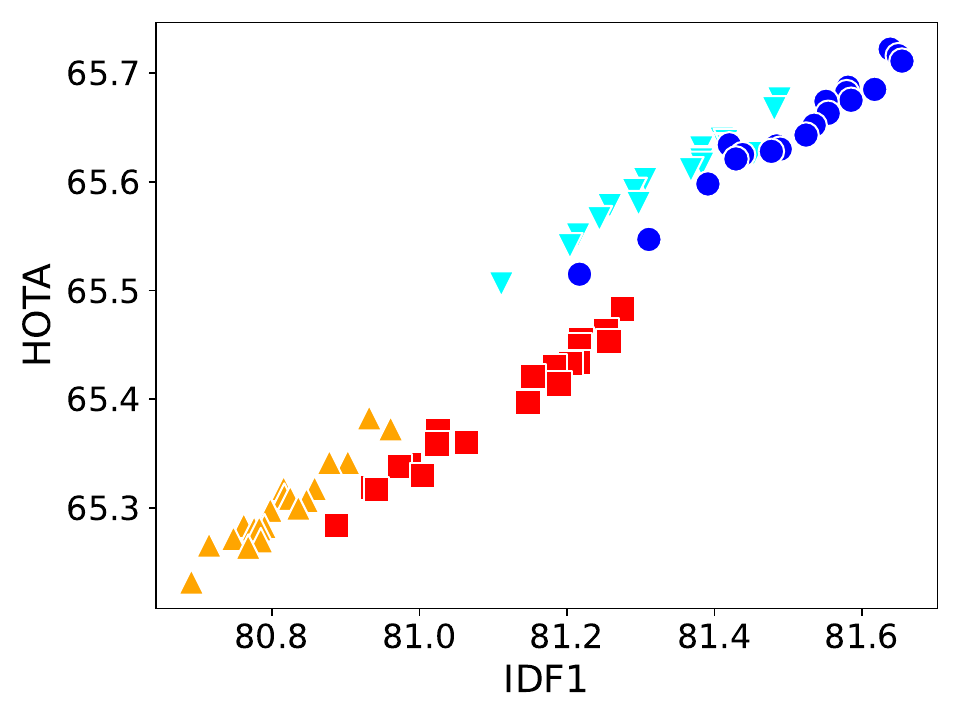} \\ 
		& \begin{subfigure}{0.05\linewidth}
		\caption{}
		\label{fig:byte_mot20}
		\end{subfigure} &
		\begin{subfigure}{0.05\linewidth}
		\caption{}
		\label{fig:botsort_mot20}
		\end{subfigure}\\	
		\rotatebox{90}{
			\parbox{0.3\textwidth}{\centering{\textbf{KITTI}}}} &  \includegraphics[scale=0.3]{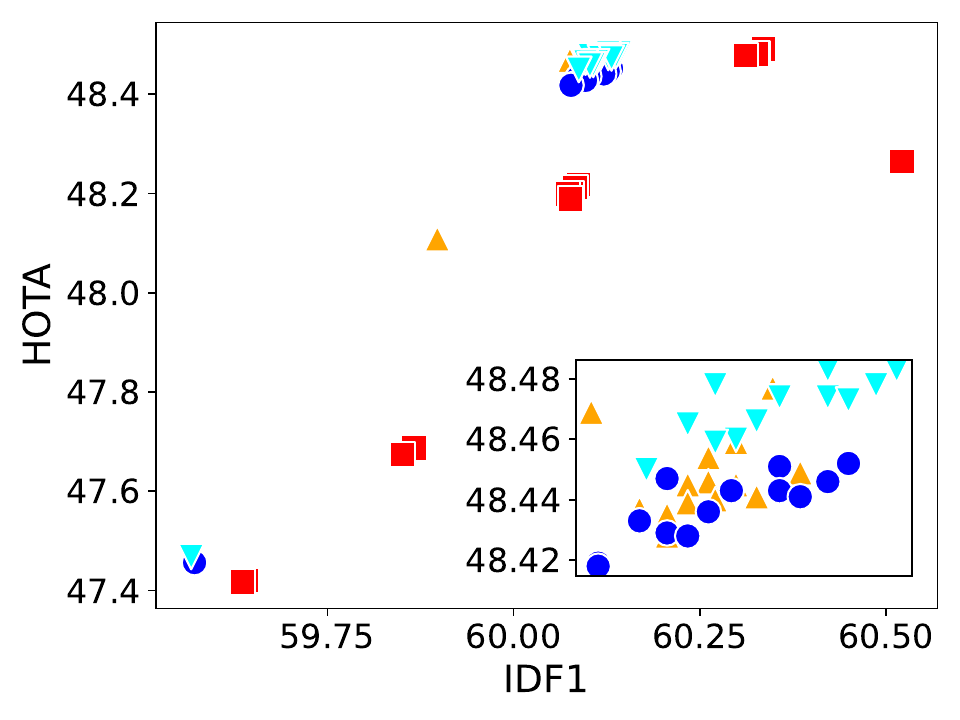}  &  \includegraphics[scale=0.3]{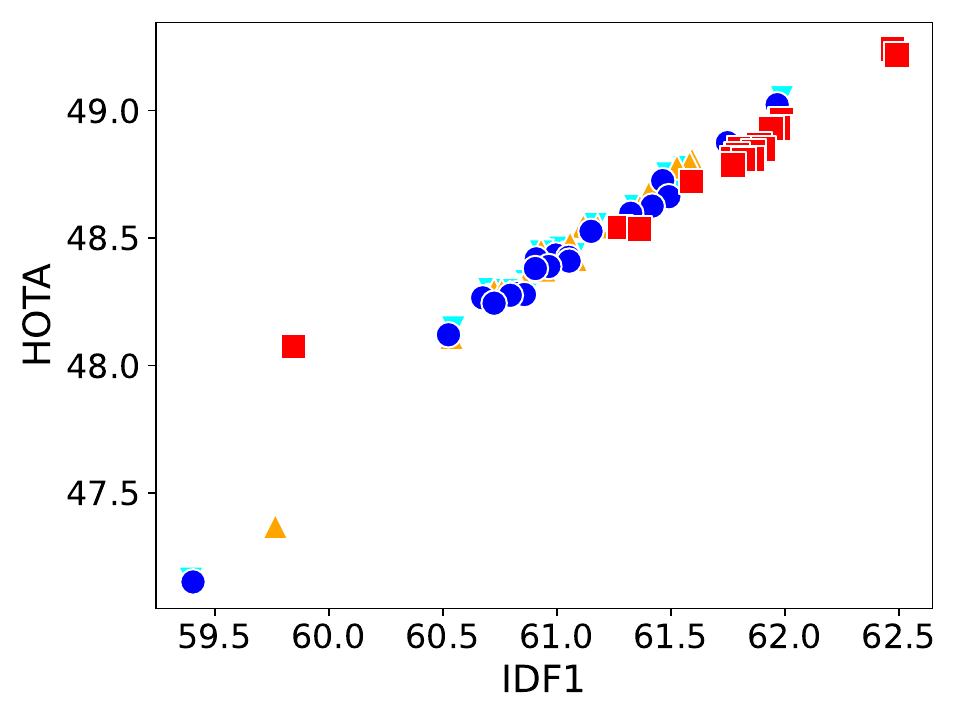} \\ 
		& \begin{subfigure}{0.43\linewidth}
		\caption{Inset zooms into top cluster.}
		\label{fig:byte_kitti}
		\end{subfigure} &
		\begin{subfigure}{0.05\linewidth}
		\caption{}
		\label{fig:botsor_kitti}
		\end{subfigure}\\	
		\rotatebox{90}{
			\parbox{0.3\textwidth}{\centering{\textbf{DanceTrack}}}} &  \includegraphics[scale=0.3]{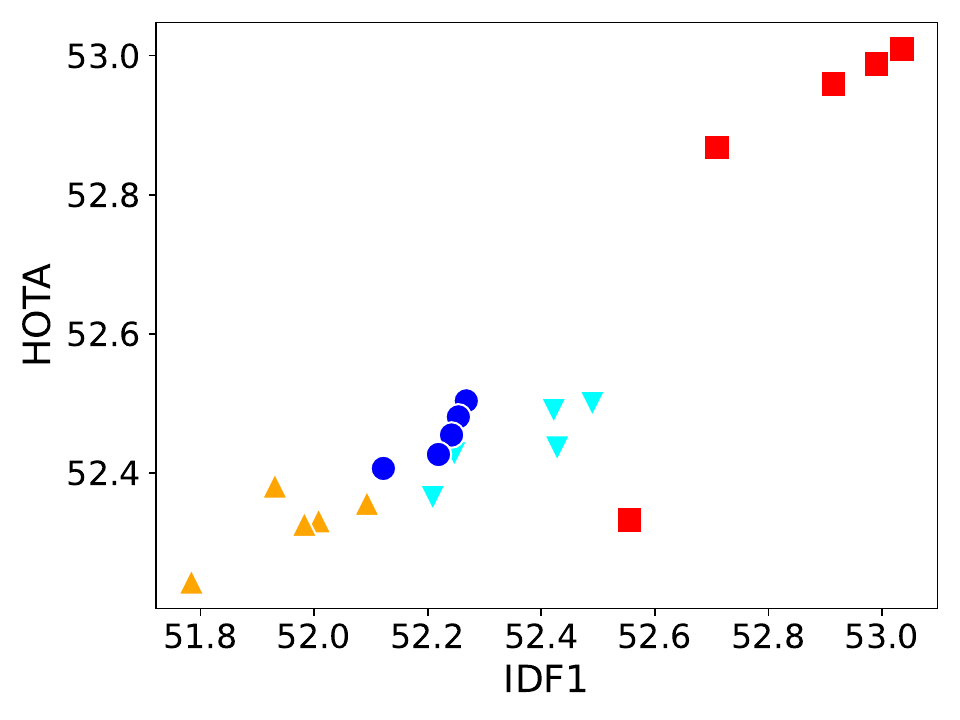}  &  \includegraphics[scale=0.3]{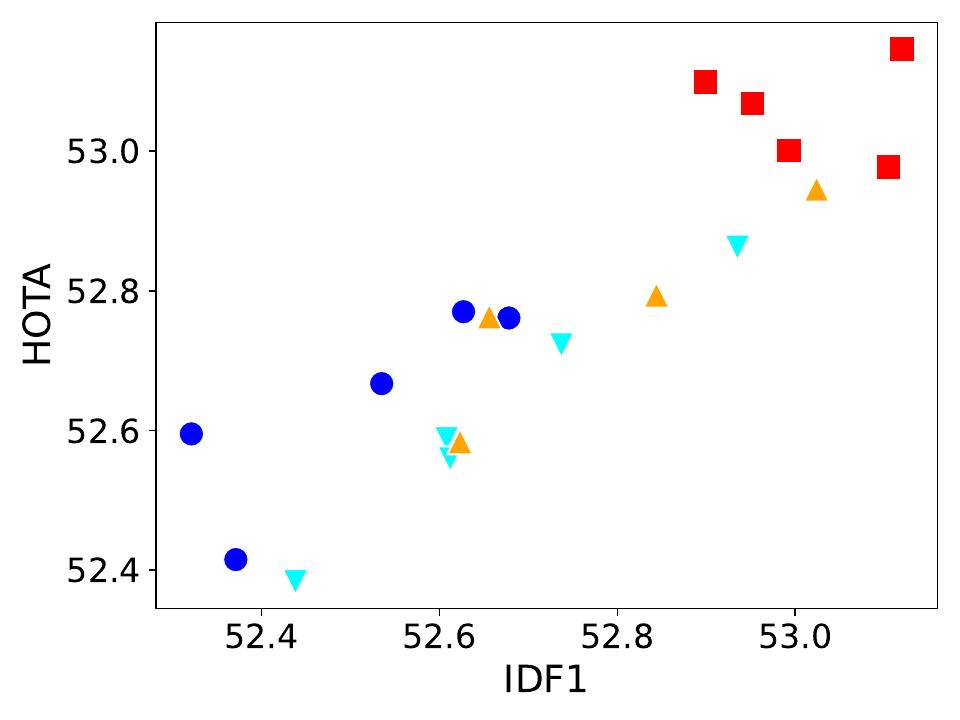} \\ 
		& \begin{subfigure}{0.05\linewidth}
		\caption{}
		\label{fig:byte_dance}
		\end{subfigure} &
		\begin{subfigure}{0.05\linewidth}
		\caption{}
		\label{fig:botsort_dance}
		\end{subfigure}\\
	\end{tabular}
	\caption{Effect, across different datasets, of modifying ByteTrack (baseline) with our proposed methods for exploiting object detector uncertainty. BoT-SORT adds CMC.}
	\label{fig:bytetrack}
\end{figure}

\subsection{Implementation details}\label{subsec:details}
\paragraph{\textbf{Custom NMS}.} We have upgraded the official Torchvision NMS algorithm by developing a custom function with a CUDA kernel for calculating the mean and variance of overlapping bounding boxes, including their confidence scores. Packaged as a C++ and CUDA extension for PyTorch, our proposed NMS modification offers easy integration as a drop-in replacement in existing trackers. Optimized for CUDA, our method can also adapt to CPU backends. We select our custom NMS as the default method for our tracking algorithms.\\

\paragraph{\textbf{Fast GMC}.} Global motion compensation (GMC) is widely used to adapt to motion and perspective shifts to improve tracking algorithms. However, traditional GMC implementations, as the one in Python introduced by BoT-SORT \cite{aharon2022bot,Jocher_YOLO_by_Ultralytics_2023} or the C++ translation in SparseTrack \cite{sparsetrack}, face accessibility and dependency issues for online deployment. Moreover, they only apply the affine transformation mentioned in \cref{subsec:cmc}. Leveraging OpenCV's C++ functions, we implement Fast GMC with the option of performing either affine or the entire homography transformation.

\paragraph{\textbf{Object detection}.} It has become usual to use YOLOX \cite{Ge2021YOLOXEY} for benchmarking multi-object trackers. In order to keep up with the latest advances in real-time object detection while developing new tracking methods, we decided to switch to the Ultralytics framework \cite{Jocher_YOLO_by_Ultralytics_2023}. Specifically, we train YOLOv8-l following the standard procedure \cite{zhang2022bytetrack} for the benchmarks of interest for this work. Instead of discarding labels with bounding boxes outside the image boundaries, we clipped them during training for the respective detection models.\\

\paragraph{\textbf{Unified framework}.} We provide a unified testbed for multi-object tracking. This allows adding several cascade association mechanisms, camera motion compensation methods, IoU disambiguation methods, hyperparameter tunning, among others. Our framework offers a possibility, similar to ensembling, of running multiple experiments concurrently, with the same tracking configuration. As the experiments run and metrics are calculated, the results can be visualized in a web browser. Particularly, for the results presented next, we ran 20 seeds per tracker configuration except for DanceTrack. For the latter, we used 5 seeds due to its much larger data volume not allowing quick exploration concurrently.


\section{Results}\label{sec:results}
In this section, we examine the effect on tracking performance of the IoU disambiguation methods of \cref{subsec:disamb} (with and without CMC), of the measured $\bm{R}_t$ in the Kalman filter (\texttt{U-Kalman}) of \cref{subsec:ukalman}, of our CMC method of \cref{subsec:cmc} for high-speed camera motion, and of our addition to the cascade associations of \cref{subsec:cascade}. Finally, we compare the results and generalize the best of our observations on the validation sets to the test set of the most complex dataset.

\paragraph{\textbf{IoU disambiguation without CMC}.} We use ByteTrack as a popular baseline. \Cref{fig:bytetrack} shows the effects of considering the uncertainty in the object detections. In most cases, the disambiguation by bounding-box sizes tends to enhance the tracking performance. For MOT17, disambiguation does not have any effect, as shown in \cref{fig:byte_mot17}. Instead, updating the Kalman filter with the measured uncertainty tends to have the most prominent role in most datasets. As expected, phase disambiguation fails for MOT20, since it is not possible to measure phase oscillations with such crowded scenes. Aditionally, the NMS procedure itself might lead to inaccurate variance estimations due to the high occlusion density. This may explain why the baseline tends to perform better in this dataset, as shown in \cref{fig:byte_mot20}.

\paragraph{\textbf{IoU disambiguation with CMC}.} Since BoT-SORT without Re-ID is effectively ByteTrack $+$ CMC with affine transformation, we use it as baseline. Comparing \cref{fig:byte_mot17} with \cref{fig:botsort_mot17}, we observe that both disambiguation methods outperform the baseline on MOT17. This is reasonable, since both methods rely on IoU, which may be zero if a ground truth has significantly shifted from a prediction due to the camera motion. Notably, by comparing the axes in both figures we observe that CMC does not boost \texttt{U-Kalman} on comparable amounts as the other methods for MOT17. In contrast, it does so for MOT20, as observed in \cref{fig:botsort_mot20} as the closing of the gap with the baseline in \cref{fig:byte_mot20}. The effect of the phase disambiguation is more marked for DanceTrack, as seen by comparing \cref{fig:botsort_dance} with \cref{fig:byte_dance}. It went from being the worse to being the second best method. It suggests that the prediction of phase oscillations helps in associating identities in the complex patterns of the choreographies. Finally, after adding CMC, the improvements over the baseline for KITTI are rather small for the disambiguation methods, with still \texttt{U-Kalman} playing the dominant role.

\begin{figure}[tb]
  \centering
  \begin{subfigure}{0.42\linewidth}
    \includegraphics[scale=0.3]{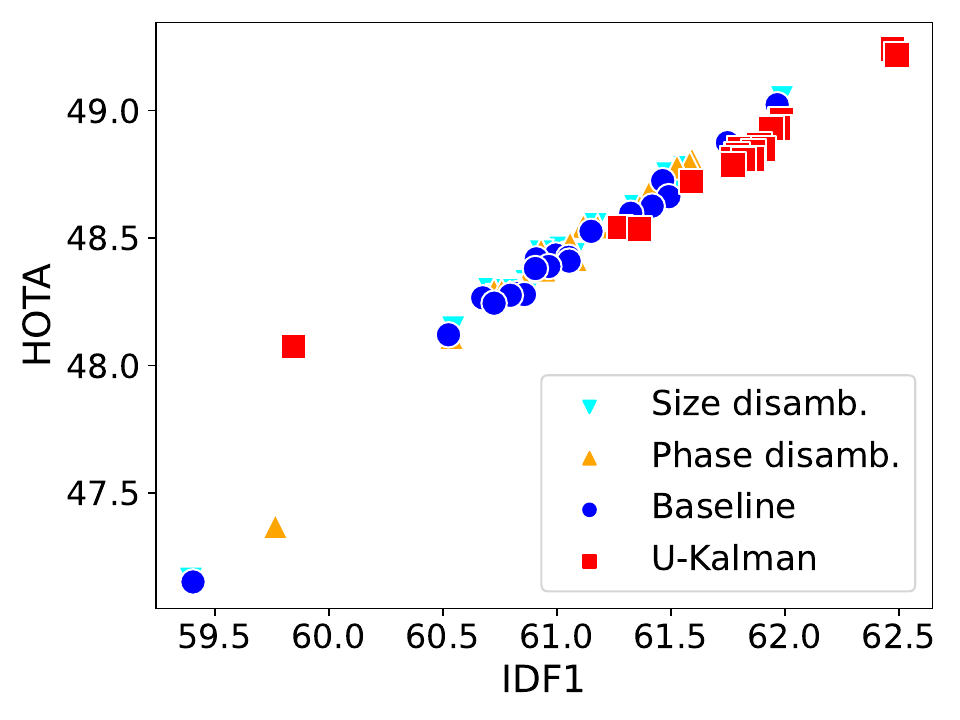} 
    \caption{BoT-SORT with affine CMC.}
    \label{fig:botsort_kitti2}
  \end{subfigure}
  \begin{subfigure}{0.42\linewidth}
    \includegraphics[scale=0.3]{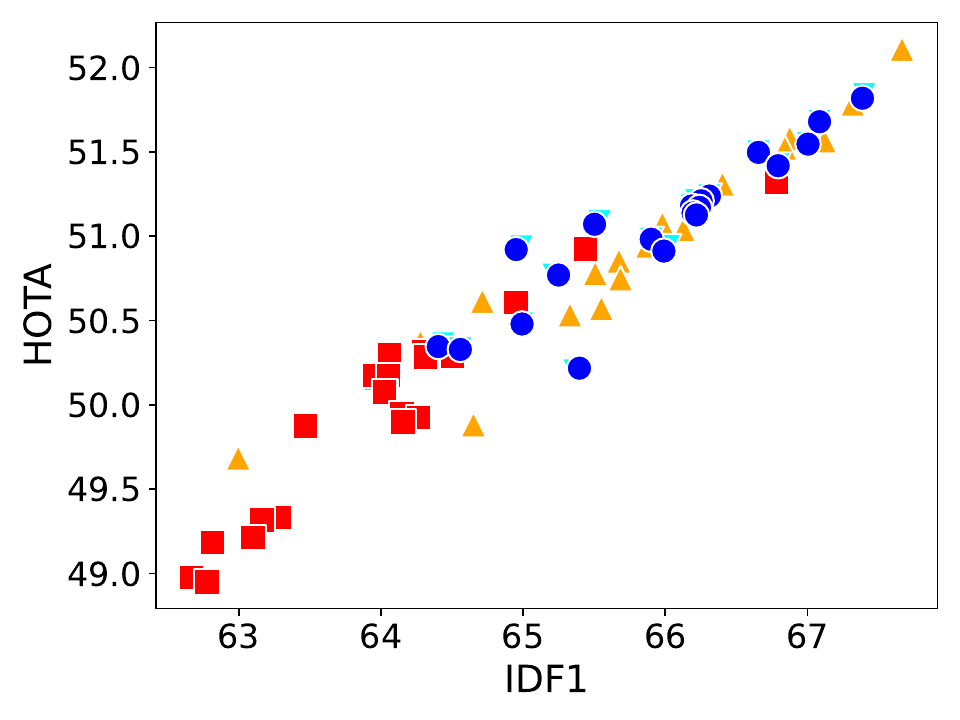}
    \caption{BoT-SORT with homographic CMC.}
    \label{fig:ubotsort_kitti}
  \end{subfigure}
  \caption{Using homography, as opposed to affine, transformation in the CMC on KITTI.}
  \label{fig:botsort_kitti_both}
\end{figure}

\paragraph{\textbf{CMC via homography transformation}.} Since KITTI exhibits large pixel displacements due to fast camera motion, it is the most suitable dataset to test our proposed method for CMC. We show the results in \cref{fig:botsort_kitti_both}. We compare with the CMC method used in BoT-SORT, which is again shown in \cref{fig:botsort_kitti2} for convenience. Our method significantly boosts in both HOTA and IDF1, as shown in \cref{fig:ubotsort_kitti}. As observed when adding CMC to ByteTrack on MOT17 (in order to get BoT-SORT), the boost to \texttt{U-Kalman} is the least. Again, the most benefited method tends to be IoU disambiguation by phase. The performance increase demonstrates the importance of considering the entire perspective transformation when dealing with scenes with fast camera motion, as may be the case with autonomous driving.

\begin{figure}[tb]
  \centering
  \begin{subfigure}{0.42\linewidth}
    \includegraphics[scale=0.3]{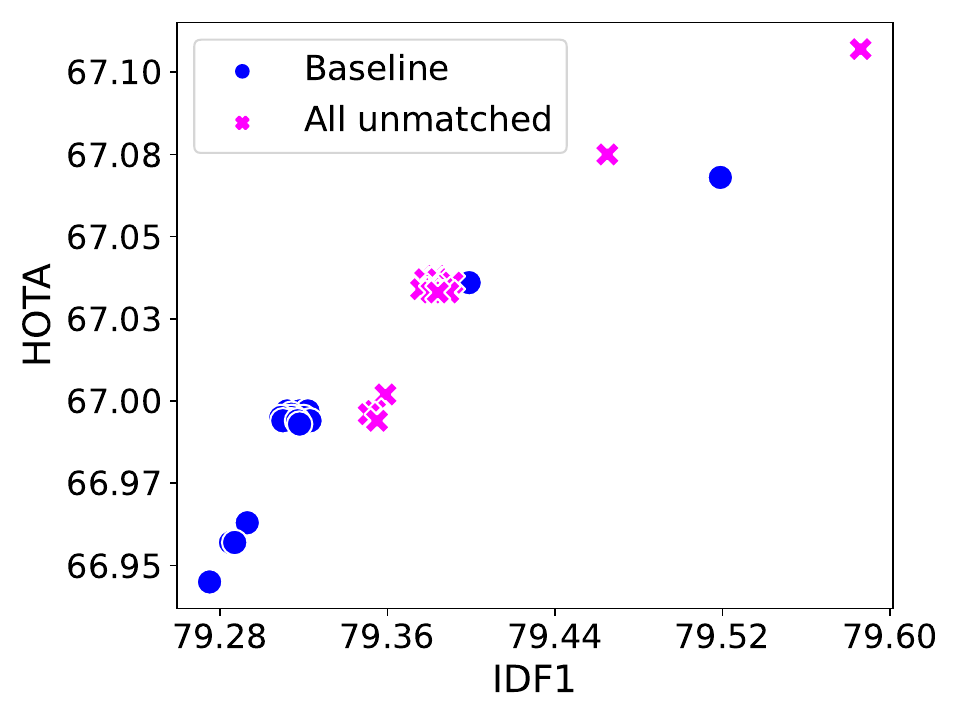} 
    \caption{SparseTrack as baseline.}
    \label{fig:deep_mot17}
  \end{subfigure}
  \begin{subfigure}{0.42\linewidth}
    \includegraphics[scale=0.3]{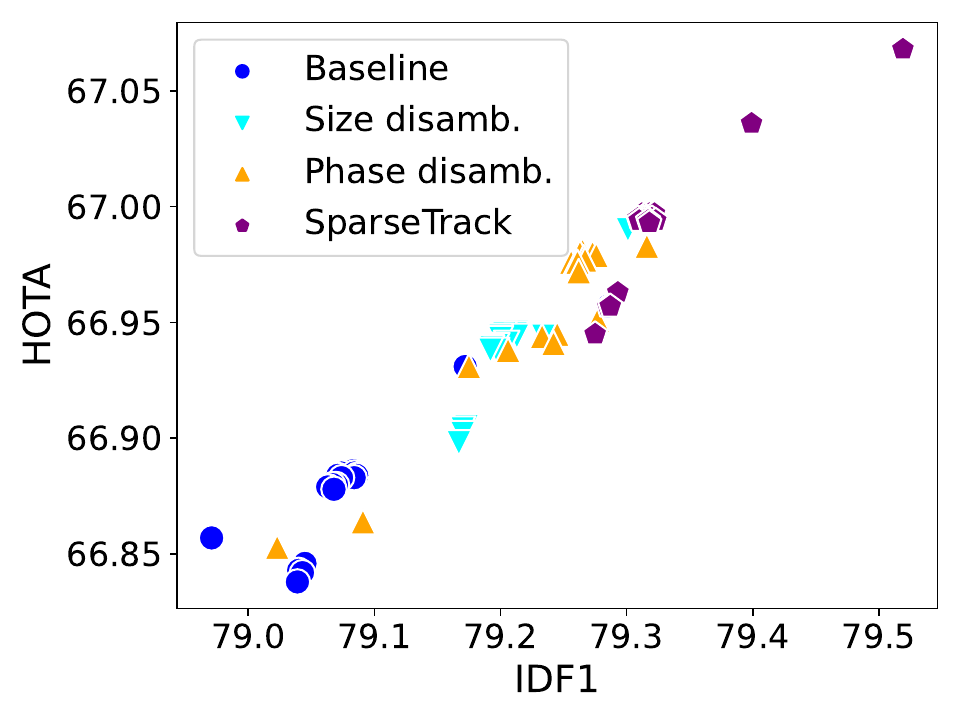}
    \caption{BoT-SORT as baseline.}
    \label{fig:mabotdeep_mot17}
  \end{subfigure}
  \caption{Effect, within MOT17, of considering all unmatched detections (from each detection bin) when confirming new tracks.}
  \label{fig:deep}
\end{figure}

\paragraph{\textbf{Varying cascade associations}.} We consider binning detections by pseudo-depth, using SparseTrack's method on MOT17. In this case, the IoU disambiguation methods bring no clear advantage --- due to pseudo-depth and bounding-box sizes being highly correlated. Since SparseTrack builds upon BoT-SORT, the lagging of \texttt{U-Kalman} observed in \cref{fig:botsort_mot17} remains. We notice a performance boost though, when considering all unmatched detections to confirm new tracks. This is shown in \cref{fig:deep_mot17}. Remarkably, when all these detections are considered in BoT-SORT, and IoU disambiguation is performed, the result overlaps with the SparseTrack baseline. This is shown in \cref{fig:mabotdeep_mot17}, and suggests ways to reach SparseTrack state-of-the art performance with fewer resources (SparseTrack has, by default, 8x more Hungarian calls in the second stage, corresponding to 8 pseudo-depth divisions of the second bin of detections).

\paragraph{\textbf{Generalization}.} We take the dataset with more data and motion complexity to generalize the observations. On average, \cref{fig:bytetrack} suggests that UTrack, in the configuration BoT-SORT + \texttt{U-Kalman}, has better generalization capabilities. We run it on the DanceTrack test set and submit the results to the designated server. \Cref{tab:dancetrack_test} shows the comparison with state-of-the-art trackers of the SORT family, \ie IoU-based methods which use Kalman filter as motion model, given observations from deep detectors. By using detection uncertainty, we observe improvements in association (AssA) over the BoT-SORT baseline ($\uparrow 5\%$). This is reflected in the main metric (HOTA), making UTrack the best among the \emph{real-time} trackers that do not train extra embeddings to capture appearance information. The overhead of considering the detection uncertainty for real-time deployment is minimal, since the result for UTrack in \cref{tab:dancetrack_test} was obtained at $\sim$27 FPS on a NVIDIA A100 GPU (compared with the $\sim$32 FPS of BoT-SORT in the same machine). Remarkably, considering the correct Kalman filter observation noise covariance, $\bm{R}_t$, gives a much stronger predictor than by using the complicated OC-SORT re-update procedure.

\begin{table}[tb]
  \caption{Results on DanceTrack-test. (*) Average between test results in their paper and in their github repository. \textbf{Bold}: best by using motion model only and no extra embeddings for appearance cues. \textit{Italics}: overall best in the SORT family.}
  \label{tab:dancetrack_test}
  \centering
  \setlength{\tabcolsep}{5pt}
  \begin{tabular}{@{}llllll@{}}
    \toprule
    Tracker & HOTA$\uparrow$ & DetA$\uparrow$ & AssA$\uparrow$ & MOTA$\uparrow$ & IDF1$\uparrow$ \\
    \midrule
    {\color{red}appearance \& motion:} & & & & &\\
    DeepSORT \cite{wojke2017simple}& 45.6 & 71.0 & 29.7 & 87.8 & 47.9\\
    StrongSORT \cite{strongSORT}& 55.6 & 80.7& 38.6 & 91.1 & 55.2 \\
    Deep OC-SORT \cite{maggiolino2023deep} & \textit{61.3} & \textit{82.2}&\textit{45.8} & \textit{92.3} & \textit{61.5} \\
    \hline
    {\color{red}motion only:} & & & & &\\
    SORT \cite{Bewley2016SORT}& 47.9 & 72.0 & 31.2 & 91.8 & 50.8\\  
    ByteTrack \cite{zhang2022bytetrack}& 47.3 & 71.6 &31.4 & 89.5 & 52.5 \\
    SparseTrack \cite{sparsetrack} & 55.5 & 78.9&39.1 & \textbf{91.3} & \textbf{58.3} \\
    OC-SORT* \cite{cao2023observation} & 55.1 & \textbf{80.4}&39.2 & 90.8 & 54.6 \\
    BoT-SORT \cite{aharon2022bot}& 53.8 & 77.8 & 37.3 & 89.7 & 56.1 \\
     UTrack (ours) & \textbf{55.8} & 79.4 & \textbf{39.3} & 89.7 & 56.4 \\
  \bottomrule
  \end{tabular}
\end{table}

\section{Conclusion}
We have shown that, by measuring the aleatoric uncertainty of a popular object detector, and propagating this information through the association procedure of simple yet powerful real-time trackers, gives noticeable gains in tracking performance. Our results suggest interesting avenues for reliable visual perception in safety-critical applications. Future research includes a better understanding of the behavior of Kalman filters with measured observation noise under camera motion compensation, as well as how the measured uncertainty affects more complex multi-object trackers. Furthermore, a comprehensive comparison with trackers using probabilistic detectors is needed, in order to establish baselines that the community can use to expand the field.\\

\noindent \textbf{Acknowledgements}\\
The authors are indebted to the reviewers for useful suggestions regarding the presentation of this work.

%
%
\bibliographystyle{splncs04}
\bibliography{main}

\begin{thebibliography}{10}
\providecommand{\url}[1]{\texttt{#1}}
\providecommand{\urlprefix}{URL }
\providecommand{\doi}[1]{https://doi.org/#1}

\bibitem{ABDAR2021243}
Abdar, M., Pourpanah, F., Hussain, S., Rezazadegan, D., Liu, L., Ghavamzadeh,
  M., Fieguth, P., Cao, X., Khosravi, A., Acharya, U.R., Makarenkov, V.,
  Nahavandi, S.: A review of uncertainty quantification in deep learning:
  Techniques, applications and challenges. Information Fusion  \textbf{76},
  243--297 (2021). \doi{https://doi.org/10.1016/j.inffus.2021.05.008},
  \url{https://www.sciencedirect.com/science/article/pii/S1566253521001081}

\bibitem{aharon2022bot}
Aharon, N., Orfaig, R., Bobrovsky, B.Z.: Bot-sort: Robust associations
  multi-pedestrian tracking. arXiv preprint arXiv:2206.14651  (2022)

\bibitem{begoli2019need}
Begoli, E., Bhattacharya, T., Kusnezov, D.: The need for uncertainty
  quantification in machine-assisted medical decision making. Nature Machine
  Intelligence  \textbf{1}(1),  20--23 (2019)

\bibitem{bernardin2008evaluating}
Bernardin, K., Stiefelhagen, R.: Evaluating multiple object tracking
  performance: the clear mot metrics. EURASIP Journal on Image and Video
  Processing  \textbf{2008},  1--10 (2008)

\bibitem{Bewley2016SORT}
Bewley, A., Ge, Z., Ott, L., Ramos, F., Upcroft, B.: Simple online and realtime
  tracking. In: IEEE International Conference on Image Processing. Phoenix, USA
  (2016)

\bibitem{brando2020}
{Brando}, A., {Torres}, D., {Rodr\'{\i}guez-Serrano}, J.A., {Vitri\`{a}}, J.:
  Building uncertainty models on top of black-box predictive apis. IEEE Access
  \textbf{8},  121344--121356 (2020)

\bibitem{cao2023observation}
Cao, J., Pang, J., Weng, X., Khirodkar, R., Kitani, K.: Observation-centric
  sort: Rethinking sort for robust multi-object tracking. In: Proceedings of
  the IEEE/CVF Conference on Computer Vision and Pattern Recognition. pp.
  9686--9696 (2023)

\bibitem{detr}
Carion, N., Massa, F., Synnaeve, G., Usunier, N., Kirillov, A., Zagoruyko, S.:
  End-to-end object detection with transformers. CoRR  \textbf{abs/2005.12872}
  (2020), \url{https://arxiv.org/abs/2005.12872}

\bibitem{Choi_2019_ICCV}
Choi, J., Chun, D., Kim, H., Lee, H.J.: Gaussian yolov3: An accurate and fast
  object detector using localization uncertainty for autonomous driving. In:
  Proceedings of the IEEE/CVF International Conference on Computer Vision
  (ICCV) (October 2019)

\bibitem{Choi_2021_ICCV}
Choi, J., Elezi, I., Lee, H.J., Farabet, C., Alvarez, J.M.: Active learning for
  deep object detection via probabilistic modeling. In: Proceedings of the
  IEEE/CVF International Conference on Computer Vision (ICCV). pp. 10264--10273
  (October 2021)

\bibitem{d2022underspecification}
D'Amour, A., Heller, K., Moldovan, D., Adlam, B., Alipanahi, B., Beutel, A.,
  Chen, C., Deaton, J., Eisenstein, J., Hoffman, M.D., et~al.:
  Underspecification presents challenges for credibility in modern machine
  learning. Journal of Machine Learning Research  \textbf{23}(226),  1--61
  (2022)

\bibitem{mot17}
Dendorfer, P., Osep, A., Milan, A., Schindler, K., Cremers, D., Reid, I., Roth,
  S., Leal-Taix{\'e}, L.: Motchallenge: A benchmark for single-camera multiple
  target tracking. International Journal of Computer Vision  \textbf{129}(4),
  845--881 (2021). \doi{10.1007/s11263-020-01400-9}

\bibitem{mot20}
Dendorfer, P., Rezatofighi, H., Milan, A., Shi, J., Cremers, D., Reid, I.,
  Roth, S., Schindler, K., Leal-Taixé, L.: Mot20: A benchmark for multi object
  tracking in crowded scenes (2020)

\bibitem{strongSORT}
Du, Y., Zhao, Z., Song, Y., Zhao, Y., Su, F., Gong, T., Meng, H.: Strongsort:
  Make deepsort great again. IEEE Transactions on Multimedia  \textbf{25},
  8725--8737 (2023). \doi{10.1109/TMM.2023.3240881}

\bibitem{aut_driv}
{Feng}, D., {Rosenbaum}, L., {Dietmayer}, K.: Towards safe autonomous driving:
  Capture uncertainty in the deep neural network for lidar 3d vehicle
  detection. In: 21st IEEE International Conference on Intelligent
  Transportation Systems. pp. 3266--3273 (2018)

\bibitem{9525313}
Feng, D., Harakeh, A., Waslander, S.L., Dietmayer, K.: A review and comparative
  study on probabilistic object detection in autonomous driving. IEEE
  Transactions on Intelligent Transportation Systems  \textbf{23}(8),
  9961--9980 (2022). \doi{10.1109/TITS.2021.3096854}

\bibitem{gawlikowski2023survey}
Gawlikowski, J., Tassi, C.R.N., Ali, M., Lee, J., Humt, M., Feng, J., Kruspe,
  A., Triebel, R., Jung, P., Roscher, R., et~al.: A survey of uncertainty in
  deep neural networks. Artificial Intelligence Review  \textbf{56}(Suppl 1),
  1513--1589 (2023)

\bibitem{Ge2021YOLOXEY}
Ge, Z., Liu, S., Wang, F., Li, Z., Sun, J.: Yolox: Exceeding yolo series in
  2021. ArXiv  \textbf{abs/2107.08430} (2021),
  \url{https://api.semanticscholar.org/CorpusID:236088010}

\bibitem{Geiger2013IJRR}
Geiger, A., Lenz, P., Stiller, C., Urtasun, R.: Vision meets robotics: The
  kitti dataset. International Journal of Robotics Research (IJRR)  (2013)

\bibitem{rev_interpret}
{Gilpin}, L.H., et~al.: Explaining explanations: An overview of
  interpretability of machine learning. In: IEEE 5th International Conference
  on Data Science and Advanced Analytics (DSAA). pp. 80--89 (2018)

\bibitem{gong2021review}
Gong, M., Wang, D., Zhao, X., Guo, H., Luo, D., Song, M.: A review of
  non-maximum suppression algorithms for deep learning target detection. In:
  Seventh Symposium on Novel Photoelectronic Detection Technology and
  Applications. vol. 11763, pp. 821--828. SPIE (2021)

\bibitem{surv_explain}
Guidotti, R., et~al.: A survey of methods for explaining black box models. ACM
  Computing Surveys  \textbf{51}(5) (2018)

\bibitem{guo2017calibration}
Guo, C., Pleiss, G., Sun, Y., Weinberger, K.Q.: On calibration of modern neural
  networks. In: International conference on machine learning. pp. 1321--1330.
  PMLR (2017)

\bibitem{hafner2019}
Hafner, D., et~al.: Reliable uncertainty estimates in deep neural networks
  using noise contrastive priors. In: Uncertainty in Artificial Intelligence
  (UAI) (2019)

\bibitem{rev_obj_det}
{Jiao}, L., {Zhang}, F., {Liu}, F., {Yang}, S., {Li}, L., {Feng}, Z., {Qu}, R.:
  A survey of deep learning-based object detection. IEEE Access  \textbf{7},
  128837--128868 (2019)

\bibitem{Jocher_YOLO_by_Ultralytics_2023}
Jocher, G., Chaurasia, A., Qiu, J.: {YOLOv8 by Ultralytics} (Jan 2023),
  \url{https://github.com/ultralytics/ultralytics}

\bibitem{khodarahmi2023review}
Khodarahmi, M., Maihami, V.: A review on kalman filter models. Archives of
  Computational Methods in Engineering  \textbf{30}(1),  727--747 (2023)

\bibitem{8917494}
Kraus, F., Dietmayer, K.: Uncertainty estimation in one-stage object detection.
  In: 2019 IEEE Intelligent Transportation Systems Conference (ITSC). pp.
  53--60 (2019). \doi{10.1109/ITSC.2019.8917494}

\bibitem{kuppers2020multivariate}
Kuppers, F., Kronenberger, J., Shantia, A., Haselhoff, A.: Multivariate
  confidence calibration for object detection. In: Proceedings of the IEEE/CVF
  conference on computer vision and pattern recognition workshops. pp. 326--327
  (2020)

\bibitem{safety_critical2018}
Le, M.T., Diehl, F., Brunner, T., Knoll, A.: Uncertainty estimation for deep
  neural object detectors in safety-critical applications. In: 2018 21st
  International Conference on Intelligent Transportation Systems (ITSC). pp.
  3873--3878 (2018). \doi{10.1109/ITSC.2018.8569637}

\bibitem{lee2022localization}
Lee, Y., Hwang, J.w., Kim, H.I., Yun, K., Kwon, Y., Bae, Y., Hwang, S.J.:
  Localization uncertainty estimation for anchor-free object detection. In:
  European Conference on Computer Vision. pp. 27--42. Springer (2022)

\bibitem{sparsetrack}
Liu, Z., Wang, X., Wang, C., Liu, W., Bai, X.: Sparsetrack: Multi-object
  tracking by performing scene decomposition based on pseudo-depth. arXiv
  preprint arXiv:2306.05238  (2023)

\bibitem{9001195}
Loquercio, A., Segu, M., Scaramuzza, D.: A general framework for uncertainty
  estimation in deep learning. IEEE Robotics and Automation Letters
  \textbf{5}(2),  3153--3160 (2020). \doi{10.1109/LRA.2020.2974682}

\bibitem{trackeval}
Luiten, J., Osep, A., Dendorfer, P., Torr, P., Geiger, A., Leal-Taix{\'e}, L.,
  Leibe, B.: Hota: A higher order metric for evaluating multi-object tracking.
  International Journal of Computer Vision pp. 1--31 (2020)

\bibitem{Maddox}
{Maddox}, W., {Garipov}, T., {Izmailov}, P., {Vetrov}, D., {Wilson}, A.G.: A
  simple baseline for bayesian uncertainty in deep learning. In: Advances in
  Neural Information Processing Systems (2019)

\bibitem{maggiolino2023deep}
Maggiolino, G., Ahmad, A., Cao, J., Kitani, K.: Deep oc-sort: Multi-pedestrian
  tracking by adaptive re-identification. In: 2023 IEEE International
  Conference on Image Processing (ICIP). pp. 3025--3029. IEEE (2023)

\bibitem{Malinin}
{Malinin}, A., {Gales}, M.: Predictive uncertainty estimation via prior
  networks. In: Proceedings of the 32nd Conference on Neural Information
  Processing Systems (2018)

\bibitem{mills2007dynamic}
Mills-Tettey, G.A., Stentz, A., Dias, M.B.: The dynamic hungarian algorithm for
  the assignment problem with changing costs. Robotics Institute, Pittsburgh,
  PA, Tech. Rep. CMU-RI-TR-07-27  (2007)

\bibitem{oksuz2023towards}
Oksuz, K., Joy, T., Dokania, P.K.: Towards building self-aware object detectors
  via reliable uncertainty quantification and calibration. In: Proceedings of
  the IEEE/CVF Conference on Computer Vision and Pattern Recognition. pp.
  9263--9274 (2023)

\bibitem{Ovadia}
Ovadia, Y., et~al.: Can you trust your model's uncertainty? evaluating
  predictive uncertainty under dataset shift. In: Advances in Neural
  Information Processing Systems 32, pp. 13991--14002 (2019)

\bibitem{pearce}
Pearce, T., et~al.: High-quality prediction intervals for deep learning: A
  distribution-free, ensembled approach. In: Proceedings of the 35th
  International Conference on Machine Learning (2018)

\bibitem{pearce2020uncertainty}
Pearce, T., Leibfried, F., Brintrup, A.: Uncertainty in neural networks:
  Approximately bayesian ensembling. In: International conference on artificial
  intelligence and statistics. pp. 234--244. PMLR (2020)

\bibitem{Picard2021Torchmanual_seed3407IA}
Picard, D.: Torch.manual\_seed(3407) is all you need: On the influence of
  random seeds in deep learning architectures for computer vision. ArXiv
  \textbf{abs/2109.08203} (2021),
  \url{https://api.semanticscholar.org/CorpusID:237563026}

\bibitem{rezatofighi2019generalized}
Rezatofighi, H., Tsoi, N., Gwak, J., Sadeghian, A., Reid, I., Savarese, S.:
  Generalized intersection over union: A metric and a loss for bounding box
  regression. In: Proceedings of the IEEE/CVF conference on computer vision and
  pattern recognition. pp. 658--666 (2019)

\bibitem{Ristani2016PerformanceMA}
Ristani, E., Solera, F., Zou, R.S., Cucchiara, R., Tomasi, C.: Performance
  measures and a data set for multi-target, multi-camera tracking. In: ECCV
  Workshops (2016), \url{https://api.semanticscholar.org/CorpusID:5584770}

\bibitem{Romano_2019}
Romano, Y., Patterson, E., Candes, E.: Conformalized quantile regression. In:
  Advances in Neural Information Processing Systems. vol.~32 (2019)

\bibitem{Rudin2019}
{Rudin}, C.: {Stop explaining black box machine learning models for high stakes
  decisions and use interpretable models instead}. Nature Machine Intelligence
  \textbf{1}(5),  206--215 (2019)

\bibitem{Sensoy}
Sensoy, M., Kaplan, L., Kandemir, M.: Evidential deep learning to quantify
  classification uncertainty. In: Advances in Neural Information Processing
  Systems. pp. 3179--3189 (2018)

\bibitem{SEONI2023107441}
Seoni, S., Jahmunah, V., Salvi, M., Barua, P.D., Molinari, F., Acharya, U.R.:
  Application of uncertainty quantification to artificial intelligence in
  healthcare: A review of last decade (2013–2023). Computers in Biology and
  Medicine  \textbf{165},  107441 (2023).
  \doi{https://doi.org/10.1016/j.compbiomed.2023.107441},
  \url{https://www.sciencedirect.com/science/article/pii/S001048252300906X}

\bibitem{SolanoPIM2021}
Solano-Carrillo, E.: Can a single neuron learn predictive uncertainty? Int. J.
  Uncertain. Fuzziness Knowl. Based Syst.  \textbf{31},  471--495 (2021),
  \url{https://api.semanticscholar.org/CorpusID:258236241}

\bibitem{pixel-attack}
Su, J., Vargas, D.V., Sakurai, K.: One pixel attack for fooling deep neural
  networks. IEEE Transactions on Evolutionary Computation  \textbf{23}(5),
  828--841 (2019). \doi{10.1109/TEVC.2019.2890858}

\bibitem{dancetrack}
Sun, P., Cao, J., Jiang, Y., Yuan, Z., Bai, S., Kitani, K., Luo, P.:
  Dancetrack: Multi-object tracking in uniform appearance and diverse motion.
  In: Proceedings of the IEEE/CVF Conference on Computer Vision and Pattern
  Recognition (CVPR). pp. 20993--21002 (June 2022)

\bibitem{SQR}
Tagasovska, N., Lopez-Paz, D.: Single-model uncertainties for deep learning.
  In: Advances in Neural Information Processing Systems 32, pp. 6414--6425.
  Curran Associates, Inc. (2019)

\bibitem{vadera2022uncertainty}
Vadera, M.P., Samplawski, C., Marlin, B.M.: Uncertainty quantification using
  query-based object detectors. In: European Conference on Computer Vision. pp.
  78--93. Springer (2022)

\bibitem{varshney2017safety}
Varshney, K.R., Alemzadeh, H.: On the safety of machine learning:
  Cyber-physical systems, decision sciences, and data products. Big data
  \textbf{5}(3),  246--255 (2017)

\bibitem{wang2020towards}
Wang, Z., Zheng, L., Liu, Y., Li, Y., Wang, S.: Towards real-time multi-object
  tracking. In: European Conference on Computer Vision. pp. 107--122. Springer
  (2020)

\bibitem{white2019homography}
White, J.H., Beard, R.W.: The homography as a state transformation between
  frames in visual multi-target tracking  (2019)

\bibitem{wojke2017simple}
Wojke, N., Bewley, A., Paulus, D.: Simple online and realtime tracking with a
  deep association metric. In: 2017 IEEE international conference on image
  processing (ICIP). pp. 3645--3649. IEEE (2017)

\bibitem{yi2024ucmctrack}
Yi, K., Luo, K., Luo, X., Huang, J., Wu, H., Hu, R., Hao, W.: Ucmctrack:
  Multi-object tracking with uniform camera motion compensation (2024)

\bibitem{YU2022104814}
Yu, J., Wang, D., Zheng, M.: Uncertainty quantification: Can we trust
  artificial intelligence in drug discovery? iScience  \textbf{25}(8),  104814
  (2022). \doi{https://doi.org/10.1016/j.isci.2022.104814},
  \url{https://www.sciencedirect.com/science/article/pii/S2589004222010860}

\bibitem{zhang2022bytetrack}
Zhang, Y., Sun, P., Jiang, Y., Yu, D., Weng, F., Yuan, Z., Luo, P., Liu, W.,
  Wang, X.: Bytetrack: Multi-object tracking by associating every detection
  box. In: European Conference on Computer Vision. pp. 1--21. Springer (2022)

\bibitem{10028728}
Zou, Z., Chen, K., Shi, Z., Guo, Y., Ye, J.: Object detection in 20 years: A
  survey. Proceedings of the IEEE  \textbf{111}(3),  257--276 (2023).
  \doi{10.1109/JPROC.2023.3238524}

\end{thebibliography}
\end{document}